\documentclass[acmsmall]{acmart}
\setcopyright{acmcopyright}
\acmJournal{PACMHCI}
\acmYear{2021} \acmVolume{5} \acmNumber{CHIPLAY} \acmArticle{231} \acmMonth{9} \acmPrice{15.00}\acmDOI{10.1145/3474658}

\AtBeginDocument{%
  \providecommand\BibTeX{{%
    \normalfont B\kern-0.5em{\scshape i\kern-0.25em b}\kern-0.8em\TeX}}}
\usepackage{verbatim}
\usepackage{graphicx}
\usepackage{algorithm}
\usepackage[noend]{algpseudocode}
\usepackage{float}
\usepackage{xcolor}
\usepackage{makecell}
\usepackage{multirow}
\usepackage{amsmath}
\usepackage{csquotes}
\usepackage{enumitem}
\DeclareMathOperator*{\argmax}{argmax}
\algnewcommand{\LineComment}[1]{\State \(\triangleright\) #1}
\newcolumntype{?}{!{\vrule width 2pt}}

\begin{document}

\title[Predicting Game Engagement and Difficulty Using AI Players]{Predicting Game Difficulty and Engagement\\Using AI Players}

\author{Shaghayegh Roohi}
\affiliation{%
	\institution{Aalto University}
	\city{Espoo}
	\country{Finland}
}
\email{shaghayegh.roohi@aalto.fi}

\author{Christian Guckelsberger}
\affiliation{%
	\institution{Aalto University}
	\city{Espoo}
	\country{Finland}
}
\email{christian.guckelsberger@aalto.fi}

\author{Asko Relas}
\affiliation{%
	\institution{Rovio Entertainment}
	\city{Espoo}
	\country{Finland}
}

\author{Henri Heiskanen}
\affiliation{%
	\institution{Rovio Entertainment}
	\city{Espoo}
	\country{Finland}
}

\author{Jari Takatalo}
\affiliation{%
	\institution{Rovio Entertainment}
	\city{Espoo}
	\country{Finland}
}

\author{Perttu H\"am\"al\"ainen}
\affiliation{%
	\institution{Aalto University}
	\city{Espoo}
	\country{Finland}
}
\email{perttu.hamalainen@aalto.fi}

\renewcommand{\shortauthors}{Roohi et al.}

\begin{abstract}
This paper presents a novel approach to automated playtesting for the prediction of human player behavior and experience. It has previously been demonstrated that Deep Reinforcement Learning (DRL) game-playing agents can predict both game difficulty and player engagement, operationalized as average pass and churn rates. We improve this approach by enhancing DRL with Monte Carlo Tree Search (MCTS). We also motivate an enhanced selection strategy for predictor features, based on the observation that an AI agent's best-case performance can yield stronger correlations with human data than the agent's average performance. Both additions consistently improve the prediction accuracy, and the DRL-enhanced MCTS outperforms both DRL and vanilla MCTS in the hardest levels. We conclude that player modelling via automated playtesting can benefit from combining DRL and MCTS. Moreover, it can be worthwhile to investigate a subset of repeated best AI agent runs, if AI gameplay does not yield good predictions on average.
\end{abstract}



\maketitle

\section{Introduction}

The development of a game typically involves many rounds of playtesting to analyze players' behaviors and experiences, allowing to shape the final product so that it conveys the design intentions and appeals to the target audience. The tasks involved in human playtesting are repetitive and tedious, come with high costs, and can slow down the design and development process substantially. Automated playtesting aims to alleviate these drawbacks by reducing the need for human participants \citep{gudmundsson2018human, chang2019reveal, stahlke2020artificial}. An active area of research, it combines human computer interaction (HCI), games user research, game studies and game design research with AI and computational modeling approaches. 

In this paper, we focus on the use of automated playtesting for player modelling, i.e.~\enquote{the study of computational means for the modeling of a player’s experience or behavior}~\citep[][p.~206]{yannakakis2018artificial}. 
A computational model that predicts experience and/or behavior accurately and at a reasonable computational cost can either assist game designers or partially automatize the game development process in many ways. For instance, it can inform the selection of the best from a range of existing designs, or be used in the procedural generation of optimal designs from a potentially vast space of possibilities \cite{togelius2008experiment, yannakakis2011experience, oulasvirta2019s}. 

Recently, researchers have started to embrace AI agents in automated playtesting, based on advances in deep reinforcement learning (DRL) which combines deep neural networks \citep{goodfellow2016deep} with reinforcement learning \citep{sutton2018reinforcement}. Reinforcement learning allows for an agent to use experience from interaction to learn how to behave optimally in different game states. Deep neural networks support this by enabling more complex mappings from observations of a game state to the optimal action. DRL has enabled artificial agents to play complex games based on visual observations alone \citep{mnih2015human}, and to exhibit diverse \citep{ijcai2020-466} and human-like \citep{ariyurek2019automated} gameplay. DRL agents can thus be employed in a wide range of playtesting tasks such as identifying game design defects and visual glitches, evaluating game parameter balance, and, of particular interest here, in predicting player behavior and experience \cite{guckelsberger2017predicting, ariyurek2019automated, hernandez2020metagame, demediuk2019challenging, ling2020using}.

In this paper, we introduce two extensions to a recent automated playtesting algorithm by \citet{roohi2020predicting}, who use DRL game-playing agents for player modelling, more specifically to predict player engagement and game difficulty. In particular, their algorithm combines simulated game-playing via deep reinforcement learning (DRL) agents with a meta-level player population simulation. The latter is used to model how players' churn in some game levels affects the distribution of player traits such as persistence and skill in later levels. Originating from an industry collaboration, they evaluated their algorithm on Angry Birds Dream Blast \citep{DB}, a popular free-to-play puzzle game. They operationalize engagement as level churn rate, i.e., the portion of players quitting the game or leaving it for an extended duration after trying a level. Game difficulty is operationalized as pass rate, i.e. as the probability of players completing a level. Understanding and estimating churn is especially important for free-to-play games, where revenue comes from a steady stream of in-game purchases and advertisement; hence, high player churn can have a strong negative impact on the game's financial feasibility. Well-tuned game difficulty is essential in facilitating flow experiences and intrinsic motivation \cite{sweetser2005gameflow, ryan2006motivational, schell2008art, denisova2017challenge}. These two factors are not independent: previous work has linked excessive difficulty to player churn \cite{bonometti2019modelling, reguera2019physics, rothmeier2020prediction}.

Roohi et al.'s algorithm strongly appeals to games industry applications by being able to model individual player differences without the costly training of multiple DRL agents. However, the pass and churn rate predictions leave room for improvement, in particular the pass rate predictions of the most difficult levels. Our two extensions overcome these shortcomings, and are informed by the following research questions: 
\begin{description}
    \item[RQ1] Considering the success of Monte Carlo Tree Search (MCTS) methods to steer AI agents in challenging games like Go \cite{silver2016mastering}, could MCTS also improve how well AI players can predict human player data?  
    \item[RQ2] Can one improve the AI gameplay features used by the population simulation, e.g., through analyzing correlations between AI gameplay statistics and the ground truth human data? 
\end{description}
To answer these questions, we have implemented and compared $2\times2\times3=12$ approaches resulting from combining the two predictive models introduced in Roohi et al.'s previous work \shortcite{roohi2020predicting} with either DRL or MCTS agents (RQ1) and three feature selection approaches (RQ2). Similar to the original paper, we apply and evaluate our extension in Angry Birds Dream Blast \citep{DB}. We find that combining DRL with MCTS, and using AI gameplay features from the best agent runs, yield more accurate predictions in particular for the pass rates. These improvements contribute to the applicability of human player engagement and difficulty prediction in both games research and industry applications.

\section{Background}
We next provide the formal basis for readers from varied disciplines to understand our contributions and related work. We first formalize game-playing, the basis of our player modelling approach, as a Markov decision process. We then introduce deep reinforcement learning and Monte Carlo tree search as the two central AI techniques in our approach.

\subsection{Markov Decision Processes}
\label{sec:mdp}
Game-playing can be considered a sequential decision-making problem. A Markov decision process (MDP)~\citep{bellman1957markovian} represents a formal framework to describe such a problem, modelling the possible interaction between an arbitrary agent and its environment over time. At each point of the interaction, the agent can receive a reward from the environment. The problem consists of finding a policy, i.e.~a set of rules associating each state with an action, that maximizes the reward which the agent can accumulate over time \citep[p.~70]{sutton2018reinforcement}.

Formally, 
the environment state at time-step $t \geq 0$ is denoted $s_t \in \mathcal{S}$. An agent's action $a_t{\in}\mathcal{A}$ can influence the future state of the environment $s_{t+1}$ determined by the environment dynamics 
given as a conditional probability distribution $p(s_{t+1}|s_t,a_t)$. The actions available to an agent can depend on the current environment state.  An action, leading to a transition between states, produces an immediate reward signal $r_{t+1} \in \mathbb{R}$, determined by the reward function 
$\rho(s_{t+1}, a_t, s_t) = r_{t+1}$.
The Markov assumption implies that $s_t$ must encode all relevant information about the past agent-environment interaction that matters for how the process unfolds in the future \citep[][pp.~1-2]{puterman2014markov}.

Here, agent and environment are given by the player and game world, respectively, and the problem consists of finding a policy to score highly in the game. The game mechanics and scoring function fulfil the role of the environment dynamics and reward function, respectively.

\subsection{(Deep) Reinforcement Learning}
\label{sec:DRL}
While an agent interacts with an MDP environment, they collect experience in the form of state, action, reward, and next state sequences $(s_t,a_t,r_{t+1},s_{t+1})$. Reinforcement learning (RL) allows an agent to solve an MDP by learning a policy $\pi(a_t|s_t)$ that maximizes the expected cumulative future reward $\mathbb{E}\left[\sum_t \gamma^t r_t\right]$ from this experience. The discount factor $\gamma \in [0, 1]$ can be used to give near-future rewards higher priority. For simple problems, the required functions such as the policy can be represented by lookup tables. However, this becomes unfeasible for problems with large state and action spaces, as present in most video games. Deep reinforcement learning (DRL) comes to the rescue by approximating these functions with deep neural networks.

In this paper, we use Proximal Policy Optimization (PPO) \cite{schulman2017proximal}, a popular, state-of-the-art DRL algorithm. It is an example of policy gradient methods, which parametrize the policy $\pi_{\theta}$ with $\theta$, the weights of a deep neural network. In the most basic of such methods, the policy is optimized based on the objective $J(\theta) = \mathbb{E} \left[ \pi_{\theta}(a_t|s_t) A(s_t,a_t)\right]$. The advantage function $A(s_t,a_t)$ captures the additional reward an agent could obtain by choosing the specific action $a_t$ over any action in the state $s_t$. Gradient ascent on this objective thus directs the policy towards actions that produce more reward in a given state. PPO contributes a modified objective function which limits the size of policy updates, allowing for smoother learning. The algorithm has been used in many applications, ranging from robot control to computer animation and game playing. However, PPO specifically and DRL more generally may require extensive multi-day training runs to learn, which is why this paper also investigates Monte Carlo tree search as an alternative.

\subsection{Monte Carlo Tree Search}\label{sec:MCTS}
Monte Carlo tree search (MCTS) is an alternative method to solving MDPs. 
It estimates the optimal action by building a tree of possible future (game) states and rewards, with each tree node corresponding to the state resulting from an explored action. The root of the tree is the current state. The algorithm iterates over four steps \cite{browne2012survey, yannakakis2018artificial}:
\begin{enumerate}
    \item \textbf{Select:} Traverse down the tree using the Upper Confidence Bound for Trees (UCT) formula until at a node with non-explored actions is hit. The selection of a child node is based on:
    \begin{equation}\label{eq:ucb}
        \argmax_{s_{d+1} \in C(s_d)} \frac{V(s_{d+1})}{N_{s_{d+1}}}+c\sqrt{\frac{\ln{N_{s_d}}}{N_{s_{d+1}}}}
    \end{equation}
    The UCT formula balances the exploration of nodes about which little is known, with the exploitation of nodes that have already shown to yield high reward. Here, $C(s_d)$ denotes the children set of the under-inspection node $s_d$ at depth $d$, and $N_{s_d}$ represents how often this node has already been visited. Moreover, $s_{d+1}$ denotes a child node, $V({s_{d+1}})$ its value, and $N_{s_{d+1}}$ how often it has been visited so far. The $c$ parameter adjusts the exploration-exploitation trade-off, and we adopt the common value \cite{kocsis2006bandit} of $\sqrt{2}$ for our experiments.
    \item \textbf{Expand:} The selected node is expanded by taking a random non-explored action. The resulting game state is added as a new child node of the selected node. 
    \item \textbf{Simulate rollout:} A forward simulation is used to advance the game from the added child with random actions until a terminal state is hit (level passed) or a move budget is consumed. The value of the added child $s_{d+1}$ at depth $d+1$ is updated to $\hat{V}(s_{d+1})=\sum_t \gamma^t r_t$, 
    and the child's visit number increases to one $N_{s_{d+1}}=1$. 
    \item \textbf{Backpropagate:} The value estimates $\hat{V}(s_{d+1})$ resulting from the simulation are propagated from the added child $s_{d+1}$ up to the root $s_0$, updating node visiting counts and values as:
    \begin{equation}
    N_{s_{d}} \gets N_{s_{d}} + 1,\quad \hat{V}(s_{d}) \gets r_{d} + \gamma \hat{V}(s_{d+1}),\quad V(s_{d}) \gets V(s_{d}) + \hat{V}(s_{d}).
    \end{equation}
\end{enumerate}
In contrast to DRL, MCTS requires no time-consuming and computationally expensive training; its estimation can be stopped at any time, and it can thus be used with arbitrary and dynamically changing computational budgets. 
MCTS is useful for long-term planning and has shown promising results in game-playing agents \cite{poromaa2017crushing, guerrero2017beyond,holmgaard2018automated}. It has gained additional popularity after being used to reach superhuman performance in the game Go \cite{silver2016mastering}.
\section{Related Work}\label{sec:related_work}
There exist numerous approaches to automated playtesting for player modelling, differing in what types of games they are applicable to, and what kind of player behavior and experience they are able to predict. We first discuss exclusively related work that employs the same state-of-the-art AI techniques that our research relies on -- see Albaghajati and Ahmed \cite{albaghajati2020video} for a more general review. We follow this up by a more specialized discussion of how difficulty and engagement relate, and how these concepts can be operationalized. To this end, we also consider computational and non-computational work beyond the previously limited, technical scope. 

\subsection{AI Playtesting With Deep Reinforcement Learning and Monte Carlo Tree Search} 
Many modern automated playtesting approaches exploit the power of deep neural networks \citep{goodfellow2016deep}, Monte Carlo tree search \citep{browne2012survey}, or a combination of these two methods. These technical approaches have emerged as dominant, as they can deal with high-dimensional sensors or controls, and generalise better across many different kinds of games. Progress in simulation-based player modelling cannot be considered separately from advancements in AI game-playing agents, and we consequently also relate to seminal work in the latter domain where it has strongly influenced the earlier.

Coming from a games industry perspective, Gudmundsson et al. \cite{gudmundsson2018human} have exploited demonstrations from human players for the supervised training of a deep learning agent to predict level difficulty in match-3 games. A deep neural network is given an observation of the game state (e.g., an array of image pixel colors) as input and trained to predict the action a human player would take in the given state. The agent's game-playing performance is used as operationalization of level difficulty. For the purpose of automatic game balancing, Pfau et al. \cite{pfau2020dungeons} similarly employ a deep neural networks to learn players' actions in a role-playing game. They analyze the balance of different character types by letting agents controlled by these neural networks play against each other. 
The drawback of such supervised deep learning methods for player modelling is that they require large quantities of data from real players, which might not be feasible to obtain for smaller companies or for a game that is not yet released.

A less data-hungry alternative is deep reinforcement learning (DRL, Section \ref{sec:DRL}), where an AI agent learns to maximize reward, e.g. game score, by drawing on its interaction experience. This makes DRL a good fit for user modeling framed as computational rationality, i.e., utility maximization limited by the brain's information-processing capabilities \cite{lewis2014computational,gershman2015computational,kangasraasio2017inferring,cheema2020predicting}. Interest in DRL for playtesting has been stimulated by Mnih et al.'s \cite{mnih2015human} seminal work on general game-playing agents, showing that a DRL agent can reach human-level performance in dozens of different games without game-specific customizations. Bergdahl et al. \cite{bergdahl2020augmenting} applied a DRL agent for player modelling to measure level difficulty, and to find bugs due to which agents get stuck or manage to reach locations in unanticipated ways. Shin et al. \cite{shin2020playtesting}  reproduce human play more time-efficiently by training an actor-critic agent \cite{mnih2016asynchronous} based on predefined strategies. Despite such advances, training a game-playing agent with DRL remains expensive: reaching human-level performance in even relatively simple games can take several days of computing time.

An alternative to facilitate automated playtesting for player modelling is to utilize a forward planning method such as MCTS (Section \ref{sec:MCTS}), simulating action outcomes to estimate the optimal action to perform.  
Several studies \cite{holmgaard2018automated, guerrero2017beyond} have engineered agents with diverse playing styles by enhancing the underlying MCTS controller with different objective functions. Holmgaard et al. \cite{holmgaard2018automated} have showed how each playing style results in different behavior and interaction with game objects in various dungeon maps. Such agents could be employed in various player modelling tasks, e.g. to evaluate which parts of a level design a human player would spend most time on. Similarly, Guerrero-Romero et al. \citep{guerrero2017beyond} employ a group of such agents to improve the design of games in the General Video Game AI (GVGAI) framework \citep{perez2016general} by analysing the visual gameplay and comparing the expected target behaviour of the individuals. Poromaa \citep{poromaa2017crushing} has employed MCTS on a match-3 game to predict average human player pass rates as a measure of level difficulty. Keehl and Smith \citep{keehl2018monster} have proposed to employ MCTS to assess human playing style and game design options in Unity games. Moreover, they propose to compare differences in AI player scores to identify game design bugs. As a last example, Ariyurek et al. \cite{ariyurek2020enhancing} have investigated the effect of MCTS modifications under two computational budgets on the human-likeliness and the bug-finding ability of AI agents. Although MCTS, in contrast to DRL, does not need expensive training, the high runtime costs restrict its application in extensive game playtesting.

It is possible to negotiate a trade-off between training and runtime cost by combining MCTS and DRL. Presently, such combinations represent the state of the art in playing highly challenging games such as Go \cite{silver2016mastering}, using learned policy and value models for traversing the tree and simulating rollouts. However, such a combination has not yet been used to drive AI agents for player behavior and experience prediction. 
This observation is captured in research question (\textbf{RQ1}), leading to the first extension to existing work contributed in this paper. 


\subsection{Operationalizing Engagement and Difficulty}
Our focus is on combining DRL and MCTS to predict pass and churn rates as measures of game difficulty and engagement, and to model the relationship between these two game metrics. This relationship has been discussed primarily from a game design and games user research point of view. Still, it remains difficult to pin down as engagement is considered one of the most complex player experiences \cite{boyle2012engagement}. Based on a review of user engagement research within and beyond video games, O'Brien and Toms define it as ``a quality of user experience characterized by attributes of challenge, positive affect, endurability, aesthetic and sensory appeal, attention, feedback, variety/novelty, interactivity, and perceived user control'' \cite{obrien2008user}, and hence identify challenge as one key influence on user engagement. Game designers often highlight the importance of carefully matching game challenge and player skill \cite{schell2008art,sweetser2005gameflow} to achieve a state of flow \cite{nakamura2014concept}, which is in turn assumed to be evoked by, and to perpetuate, high levels of engagement \cite{chen2007flow}. 
Achieving such an engaging challenge-skill trade-off is non-trivial; Lomas et al. \cite{lomas2013optimizing} for instance have studied an educational game and found that higher engagement was associated with lower game difficulty. An additional link between difficulty and engagement is postulated by self-determination theory (SDT), which explains human intrinsic motivation through the satisfaction of basic psychological needs. One of these is the feeling of competence, which is affected by the challenges experienced in performing a certain task \cite{ryan2000intrinsic}. In recent years, SDT has become increasingly popular in game research \cite{ryan2006motivational,tyack2020self}.

We account for the complexity of engagement as player experience \cite{boyle2012engagement} by operationalizing it in terms of player churn, which can be triggered by any of the underlying factors \cite{obrien2008user}. Existing computational work has considered the churn probability of an individual player \cite{bonometti2019modelling, rothmeier2020prediction}, or the average churn rate measured over a population of players and a part of the game such as a single game level \cite{reguera2019physics, roohi2020predicting}. Predicting the churn probability of a particular player allows game developers to offer personalized incentives to continue playing. Predicting the average churn rate of a game level in contrast can be used, amongst others, to detect and correct less engaging levels before they are released to the players. 

Difficulty has been operationalized based on features extracted from static game content such as gap size and the player’s maximum jump length in a platformer game \cite{sorenson2011generic}, or the board layout and pieces in puzzle games \cite{van2015automated}. While these static measures are fast to compute, they do not generalize well beyond at most narrow game genres. Moreover, they are insensitive to difficulty arising from the interaction of the player and game facets such as the level design and mechanics during play. As solution, difficulty has also been operationalized by extracing features from simulated gameplay, such as the times the player dies \citep{yannakakis2007towards} or the achieved score \cite{isaksen2017simulating, nielsen2015general}. While the earlier feature does not generalize well, a game's score, if provided at all, can be stronger affected by certain types of challenge than others, e.g. by performative, cognitive \cite{cox2012not} and decision-making \cite{denisova2020measuring} challenge, and less by e.g. emotional challenge \cite{cole2015emotional, bopp2018odd}. Crucially, while we similarly choose to extract features from simulated gameplay, operationalizing difficulty as pass rate, as similarly proposed by Poromaa \citep{poromaa2017crushing}, is sensitive to all these challenge types and agnostic with respect to the game having a score. It thus has a higher potential to be applicable across many games.


Kristensen et al. \cite{kristensen2020estimating} predict difficulty in a match-3 game based on metrics extracted from the performance of RL game-playing agents. Importantly for us, they hypothesize that performance metrics derived from an AI agent's best attempts in playing a game level, assessed over multiple runs, can better predict the perceived difficulty of human players than metrics based on average agent performance. Their hypothesis rests on the observation that, in contrast to their human ground truth data, the number of moves required by AI agents to complete a level follows a long-tailed distribution. This is likely caused by the availability of an unlimited move budget and the stochasticity of their testbed, a property shared by many games. They propose to tighten the distribution, thus moving it closer to the human ground truth, by using the best, i.e., shorter, runs only. In their specific study, they found that the number of moves a DRL agent needs for completing a level has the highest correlation with human player pass rates if one uses the maximum found among the top 5\% best runs. 
We test their hypothesis by modifying the AI gameplay features used in Roohi et al.'s \shortcite{roohi2020predicting} original approach (see \textbf{RQ2}). This constitutes a part of the second extension contributed in this paper. 

\section{Original Method}\label{sec:primary-article-details}
In their original approach, Roohi et al. \shortcite{roohi2020predicting} trained a DRL agent using the Proximal Policy Optimization (PPO, Section \ref{sec:DRL}) algorithm \cite{schulman2017proximal} on each level of Angry Birds Dream Blast \cite{DB}, which is a commercial, physics-based, non-deterministic, match-3, free-to-play mobile game. Below, we briefly summarize their approach to provide the required background for understanding our extensions. We briefly outline the details of their approach that are relevant for our extension.

\subsection{AI Problem Formulation}
To facilitate the DRL (Section \ref{sec:DRL}) training, Roohi et al. \cite{roohi2020predicting} adopted the popular game AI practice of formulating their game as a Markov Decision Process (MDP, Section \ref{sec:mdp}). The state space results from combining a $84 \times 84 \times 3$ RGB pixel array rendering of the game, capturing what the player can observe during play, with a numerical vector, representing game state details such as the goals left to achieve. The action space includes $32\times32$ possible positions in a discretized game screen to select the closest available match or booster. The rewards include a small penalty for using moves and a larger penalty for losing. Moreover, they comprise bonuses for winning, clearing goals, unlocking locks, and advancing spatially through the level, amongst others. In addition to the extrinsic rewards coming from the environment, they let the agent compute an intrinsic curiosity reward \cite{pathak2017curiosity} to facilitate efficient exploration. 

They parametrize their DRL with an experience budget of 10240 and a batch size of 1024. For each level, they provide the agent with 4 times the moves available to human players in the same level. Agent experience is collected over a time horizon of 1024. Further hyperparameters are given by a decaying learning rate of 0.0003, a discount factor of 0.99, a generalized advantage estimation parameter of 0.95, a PPO clipping parameter of 0.2, a PPO entropy coefficient of 0.005, and a curiosity reward coefficient of 0.02.

\subsection{Data Collection}
Roohi et al. \cite{roohi2020predicting} use the trained DRL agent to extract 16 features from simulated gameplay in each level: average and standard deviations of level pass rate, as well as statistics (average, standard deviation, min, max, 5th, 10th, 25th, 50th, 75th percentiles) of both cleared level goals and moves left when passing the level.

To fit their predictive models, they moreover obtained a ground truth dataset of average pass and churn rates by 95266 human players, collected from the first 168 levels of Angry Bird Dream Blast. This is a live game which is constantly being updated and improved, and the data corresponds to a particular version that is no longer online. An individual player's pass rate as operationalization of difficulty was computed as one divided by the number of required attempts, evaluating to zero if the player never passed a level. The average churn rate as a measure of engagement was defined as the portion of players that stopped playing for at least seven days after trying the level at least once. 

\subsection{Pass and Churn Rate Prediction}
\begin{figure*}[t!]
	\centering
	\includegraphics[width=0.9\textwidth]{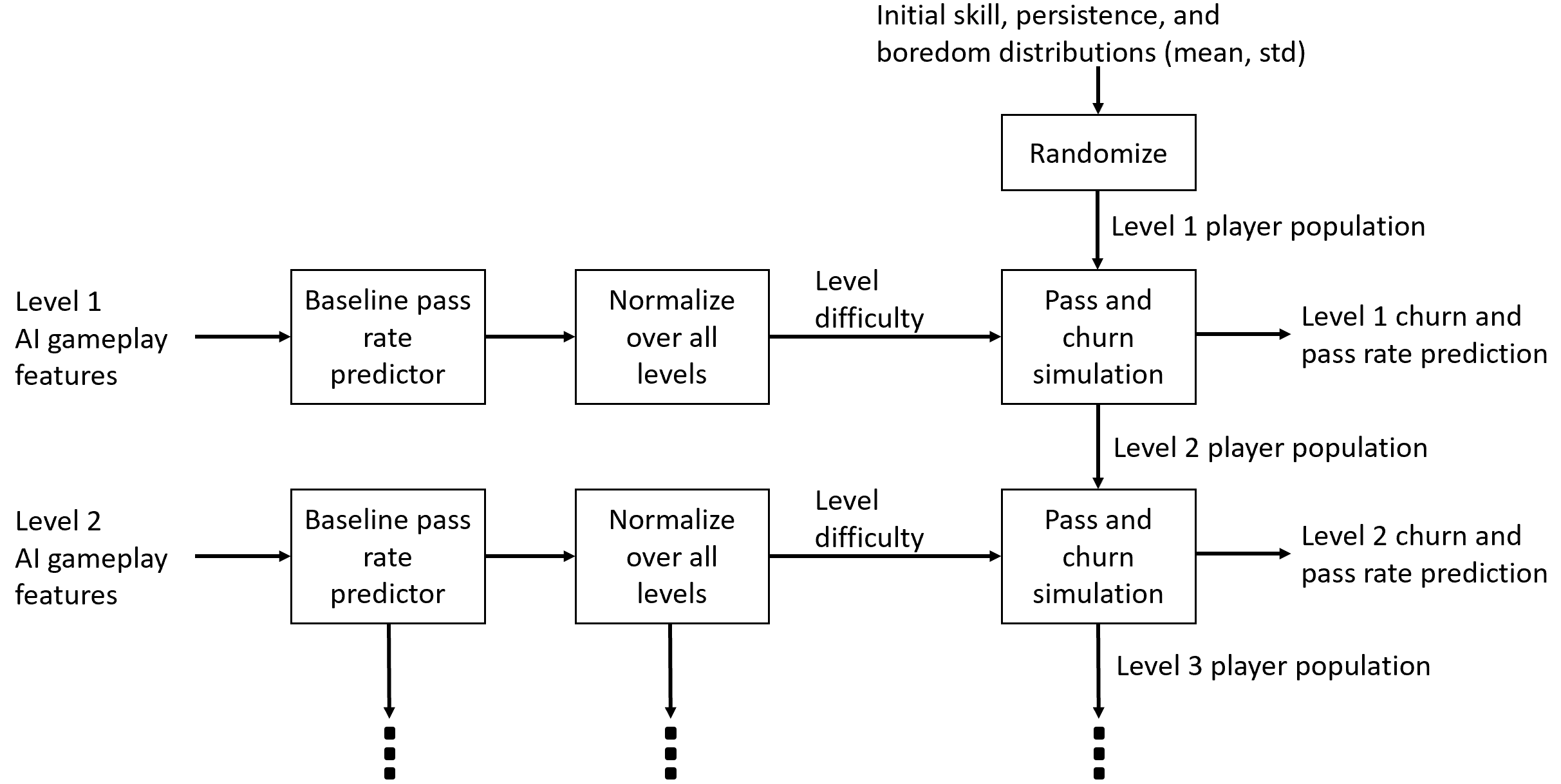}
	\caption{Overview of Roohi et al.'s \cite{roohi2020predicting} player population simulation (i.e., their \enquote{extended model}). The game level difficulties, estimated through AI gameplay features, and the player population parameters are passed to the pass and churn simulation block, which outputs the pass and churn rate of the present level and the remaining player population for the next level. Reproduced with permission.}
	\label{fig:mc_algorithm}
\end{figure*}

\begin{figure}[b!]
	\centering
	\includegraphics[width=0.99\textwidth]{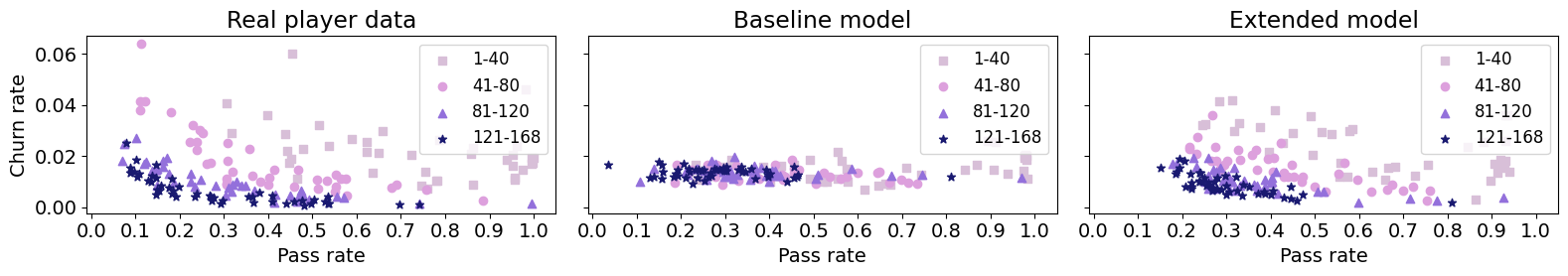}
	\caption{Roohi et al's  \cite{roohi2020predicting} original approach. Human vs. predicted pass and churn rates over four groups of 168 consecutive game levels, comparing a baseline and extended model. The colors and shapes correspond to level numbers. The baseline model produces predictions via simple linear regression. The extended model simulates player population changes over the levels, which better captures the relationship between pass and churn rates over game level progression. Reproduced with permission.}
	\label{fig:pass-churn-relation}
\end{figure}

Roohi et al. \cite{roohi2020predicting} have put forward two methods for predicting the human pass and churn rates. Their ``baseline model'' uses simple linear regression to map the input features to pass and churn rates. Their ``extended model'' employs a meta-level player population simulation (Figure \ref{fig:pass-churn-relation}), modelling individual players with the attributes skill, persistence, and the tendency to get bored. For the latter model, the baseline pass rate prediction was first normalized over the levels to produce a level difficulty estimate. This and each simulated player's skill attribute were used to determine whether the player passes a level. If a player did not pass, they were assumed to churn with a probability determined by their persistence. Players are moreover assumed to churn randomly based on their boredom tendency. In summary, the population simulation of each level uses a DRL difficulty estimate and current player population as its input, and outputs pass and churn rates for the present level, as well as the remaining player population for the next level (Figure \ref{fig:mc_algorithm}). This way, Roohi et al. were able to model how the relation of human pass and churn rates changes over the levels.

We next introduce two extensions to Roohi et al.'s original method, based on the research questions put forward in the introduction. We investigate the effect of combining DRL with MCTS to predict human players' pass and churn rates (\textbf{RQ1}). We moreover explore a different feature selection method for the pass and churn rate predictive models, based on analyzing the correlation between AI gameplay statistics and human pass rates (\textbf{RQ2}).
\section{Proposed Method}\label{sec:method}
In this paper, we extend Roohi et al.'s \cite{roohi2020predicting} original approach by complementing their DRL game play with MCTS, and by improving the features extracted from the AI gameplay data. By utilizing the same high-level population simulation, reward function, and DRL agents trained for each level, we allow for a direct comparison with their original work. 

Two observations in the original work provided initial motivation for our extensions: 1) churn predictions were more accurate if using ground truth human pass rate data instead of simulated predictions, and 2) the simulated pass rate predictions were particularly inaccurate for the most difficult levels. We hypothesize that employing MCTS in addition to DRL can improve the predictions, as the combination has been highly successful in playing very challenging games like Go \cite{silver2016mastering}. We moreover adopt Kristensen et al.'s \cite{kristensen2020estimating} hypothesis that the difficulty of the hardest levels could be better predicted by the agent's best case performance rather than average performance. To implement this, we employed multiple attempts per level and only utilized the data from the most successful attempts as gameplay features. We next describe the two resulting extensions in detail.

\subsection{Monte-Carlo Tree Search}
To use MCTS for pass and churn rate prediction, we first had to identify which variant of the algorithm allows an AI agent to successfully play even the hardest levels of our game. We elaborate on this selection first, followed by a detailed description of our agent implementation.

\subsubsection{MCTS Variants}\label{sec:MCTSvariants}
It became clear early on that MCTS with no modifications requires prohibitively many moves to pass a level and does not produce better results than Roohi et al.'s \cite{roohi2020predicting} original DRL approach. To address the problem, we determined the five hardest game levels based on the average human pass rate, compared the following MCTS variants, and selected the best for further investigation.
\begin{description}[labelindent=1em, leftmargin=3.5em]
    \item[Vanilla MCTS:] The basic algorithm described in section \ref{sec:MCTS}. This variant do not use any reward discounting ($\gamma = 1$).
    \item[DRL MCTS:] In contrast to Vanilla MCTS, this variant does not select random actions for rollout but samples them from a DRL game-playing policy, trained similarly as in Roohi et al.'s \cite{roohi2020predicting} original work. This is a generally applicable technique inspired by AlphaGo \cite{silver2016mastering}. Decreasing randomness in action selection reduces variance in the backpropagated values, which can result in better node selection and expansion. An alternative view is that MCTS improves on the results of an imperfect DRL agent. 
    \item[Myopic MCTS:]In contrast to Vanilla and DRL MCTS, this variant gives a larger weight to near-future rewards by discounting with $\gamma = 0.9$ in the rollout value computation $\hat{V}(s_{d+1})=\sum_t \gamma^t r_t$. The value estimate backpropagation to the tree root, $\hat{V}(s_{d}) \gets r_{d} + \gamma \hat{V}(s_{d+1})$, also uses discounting.
    \item[DRL-Myopic MCTS:] This version combines both previously outlined modifications. 
\end{description}

\subsubsection{MCTS Implementation Details}
Our MCTS implementations use the same moves available for human players. At each decision step, the algorithm is iterated 200 times, resulting in a tree depth of approximately 4. We assign a maximum of 10 rollout moves to each simulation step; a simulation is thus stopped either when exceeding 10 moves or when reaching a terminal state. A discount factor $\gamma=0.9$ was used for the myopic variants.

To reduce the computing cost, we assume that the game dynamics are deterministic, i.e., the same action from the same current state always results in the same next state. This allows reusing the subtree below the selected action when moving to the next game state and restarting the tree search. 
In reality, Angry Birds Dream Blast is highly random, but this is compensated by running the agent multiple times for each level, which is anyway needed to collect more reliable gameplay statistics.


\subsubsection{Selecting the Best MCTS Variant on Hard Game Levels}
\label{sec:selecting_best_mcts}
To select the best MCTS variant, we ran each candidate 20 times on each level. One run used 16 parallel instances of the game. This process took approximately 10 hours per level on a 16-core 3.4 GHz Intel Xeon CPU. 

Table \ref{tab:mctsVersions} shows the average performance of these four MCTS variants over 20 runs on the five hardest levels of the dataset. DRL-Myopic MCTS performs best. We thus select this variant for comparison against  Roohi et al.'s \cite{roohi2020predicting} original DRL-only approach, and for testing the effect of the feature extension described in Section \ref{sec:features}. 
\begin{table*}[tbp]
\centering
\caption{Performance of different MCTS variants 
on the five hardest levels in the original dataset. The Myopic variants discount future rewards, and the DRL variants sample rollout actions from a DRL policy. Overall, our DRL-Myopic MCTS outperforms the other MCTS variants and the original DRL agent of Roohi et al. \cite{roohi2020predicting}. }~\label{tab:mctsVersions}
\resizebox{\textwidth}{!}{
\begin{tabular}{|c?c|c|c|c|c?c|c|c|c|c?c|c|c|c|c|}
\hline
Method &\multicolumn{5}{c?}{Pass rate} &\multicolumn{5}{c?}{Average moves left ratio}&\multicolumn{5}{c|}{Average cleared goal percentage}\\ \cline{2-16} & 116 & 120 & 130 & 140 & 146 & 116 & 120 & 130 & 140 & 146 & 116 & 120 & 130 & 140 & 146 \\ \Xhline{4\arrayrulewidth}
Vanilla MCTS & 0 & 0 & 0.15 & 0 & 0 & 0 & 0 & 0.04 & 0 & 0 & 0.24 & 0 & 0.67 & 0.29 & 0.4 \\ \hline
DRL MCTS & 0 & 0.1 & \textbf{0.55} & 0 & 0 & 0 & 0.02 & \textbf{0.08} & 0 & 0 & 0.2 & 0.1 & \textbf{0.82} & 0.29 & 0.02 \\ \hline
Myopic MCTS & 0.2 & 0.3 & 0.05 & 0.05 & 0.05 & 0.02 & 0.03 & 0.01 & 0 & 0 & 0.56 & \textbf{0.78} & 0.72 & 0.58 & 0.76 \\ \hline
DRL-Myopic MCTS & \textbf{0.6} & \textbf{0.4} & 0.15 & \textbf{0.2} & \textbf{0.45} & \textbf{0.06} & \textbf{0.04} & 0.01 & \textbf{0.01} & \textbf{0.05} & \textbf{0.82} & 0.7 & 0.78 & \textbf{0.59} & \textbf{0.89} \\ \hline
\makecell{DRL (Roohi et al. \cite{roohi2020predicting})} & 0 & 0.002 & 0.006 & 0 & 0 & 0 & 0.0015 & 0.0046 & 0 & 0 & 0.13 & 0.43 & 0.33 & 0.31 & 0.08 \\ \hline
\end{tabular}
}
\vspace{-0.3cm}
\end{table*}

\subsection{Feature Selection: Correlating AI Gameplay Statistics With Human Data}\label{sec:features}
In addition to implementing the MCTS variants as alternatives to using DRL only, we have also tested and compared various feature selection strategies to extend Roohi et al's \cite{roohi2020predicting} original approach. The latter uses a total of 16 input features for the pass rate predictor, computed over multiple AI agent runs per level. However, this amount of features is fairly high, which may lead to the pass rate prediction model overfitting the human pass rate data. In this paper, we focus on those features that are most highly correlated with the ground truth human pass rates. We have implemented and compared the following three feature selection strategies:

\begin{description}[labelindent=1em, leftmargin=3.5em,style=nextline]
    \item [F16:] The original 16 features of Roohi et al. \cite{roohi2020predicting}.
    \item [F3:] A subset of 3 out of the 16 original features with the highest Spearman correlations with human pass rates: AI pass rate ($\rho{=}0.80$), cleared goals percentage ($\rho{=}0.60$), and the ratio of moves left to maximum allowed moves ($\rho{=}0.74$), computed as averages over 1000 runs of the DRL agent per level.
    \item [F3P:] Similar to F3, but with averages computed from a subset of the best rather than all runs, characterized by a high number of moves left after passing a level.
\end{description}

\begin{figure}[hbt!]
	\centering
	\includegraphics[width=0.5\textwidth]{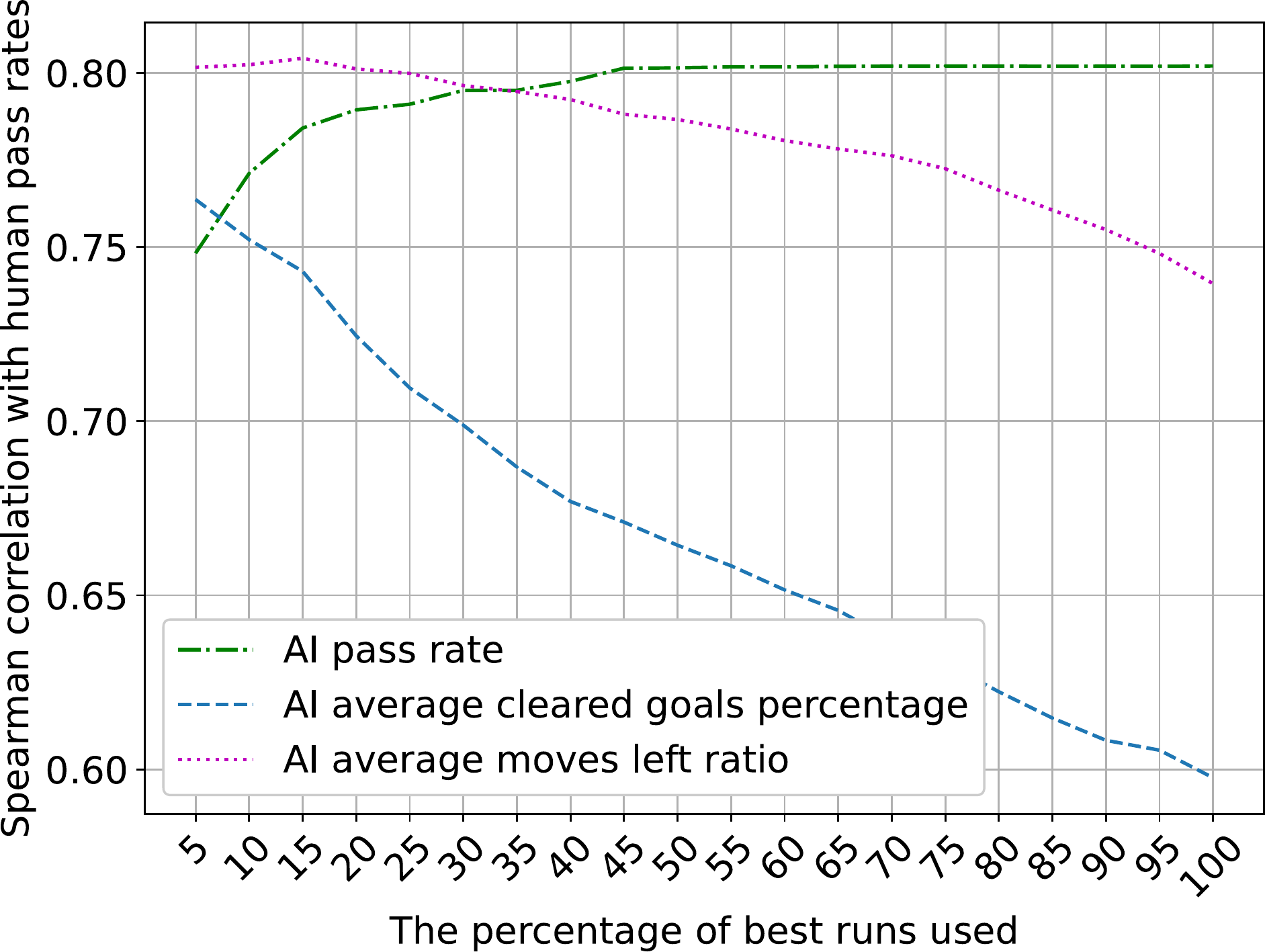}
	\caption{Spearman correlation between the F3 DRL gameplay features and human pass rates for different percentages of the best runs.}	
	\label{fig:rl_correletion}
\end{figure}

The last selection strategy, F3P, is inspired by Kristensen et al. \cite{kristensen2020estimating}, who have observed a stronger correlation between human ground truth data and features extracted from best runs (see Section \ref{sec:related_work}). To back this up and to identify the best sample size, we computed the Spearman correlation between the ground truth human pass rate and the average of the DRL agent's performance over different percentages of the agent's best runs (Figure \ref{fig:rl_correletion}). The results indicate that human pass rates are predicted best from AI pass rates if the latter are calculated over all AI runs. However, the average moves left and average cleared goals features correlate most with human pass rates if calculated from the best AI runs. This supports Kristensen et al.'s \cite{kristensen2020estimating} hypothesis for at least some, although not all, investigated gameplay features.

Based on the results in Figure \ref{fig:rl_correletion}, we used the top 15\% runs for computing the average moves left, and top 5\% for computing the average cleared goals percentage features. Pass rate was averaged over all runs. To compute the averages, we performed 20 MCTS runs and 1000 DRL runs per level. MCTS is much more computationally expensive, and the limited number of runs means that the F3P features cannot be reliably computed from MCTS data. Thus, when using F3P features with MCTS, we combine both F3 from MCTS runs and the F3P features from DRL.

\section{Results}\label{sec:evaluation}
We compare the performance of Roohi et al.'s original \cite{roohi2020predicting} DRL model and our DRL-enhanced MCTS extension, denoted as DRL-Myopic MCTS in Section \ref{sec:MCTSvariants}. We also assess the impact of the three different feature selections strategies introduced in Section \ref{sec:features}, including F16 from the original approach. To enable a fair comparison with previous work, we combining these algorithms and feature selection strategies with the two pass and churn rate predictors from Roohi et al., i.e., the baseline linear regression predictor and extended predictor with player population simulation. Taken together, this yields $2$ (algorithms) $\times\ 3$ (feature selection strategies) $\times\ 2$ (models) $=12$ combinations to evaluate. 

\begin{figure}[hbt!]
	\centering
	\includegraphics[width=0.99\textwidth]{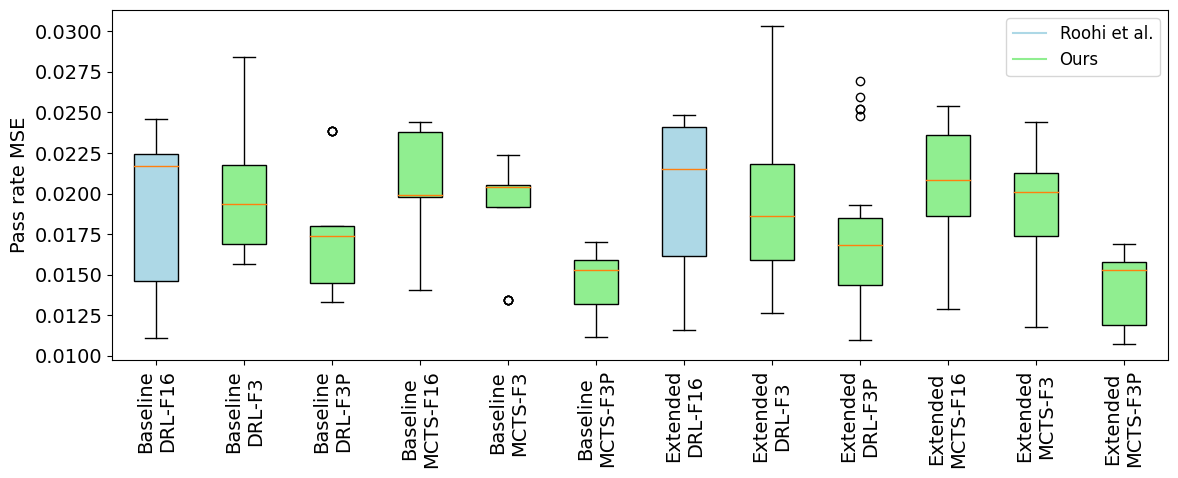}
	\medskip
	\medskip
	\includegraphics[width=0.99\textwidth]{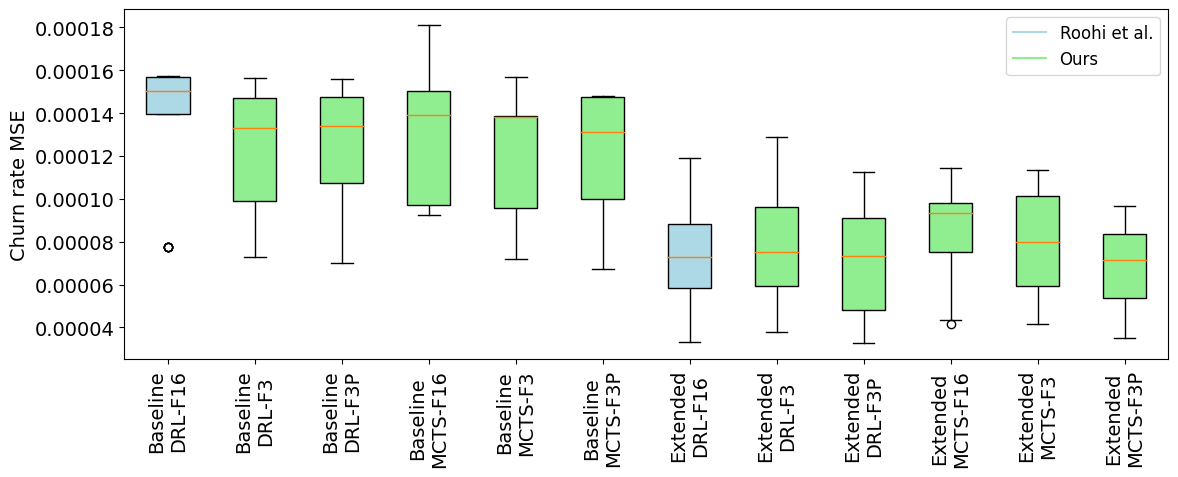}
	\caption{Boxplots of the cross-validation Mean Squared Errors for all 12 compared configurations. The configurations corresponding to the original approaches of Roohi et al. \cite{roohi2020predicting} are shown in blue. Our proposed F3P features consistently outperform the other features in pass rate predictions, and our Extended-MCTS-F3P configuration produces the overall most accurate predictions.}	
	\label{fig:passrate_mses_boxplots}
	\medskip
	\medskip
	\medskip
	\medskip
	\medskip
	\centering
	\includegraphics[width=0.99\textwidth]{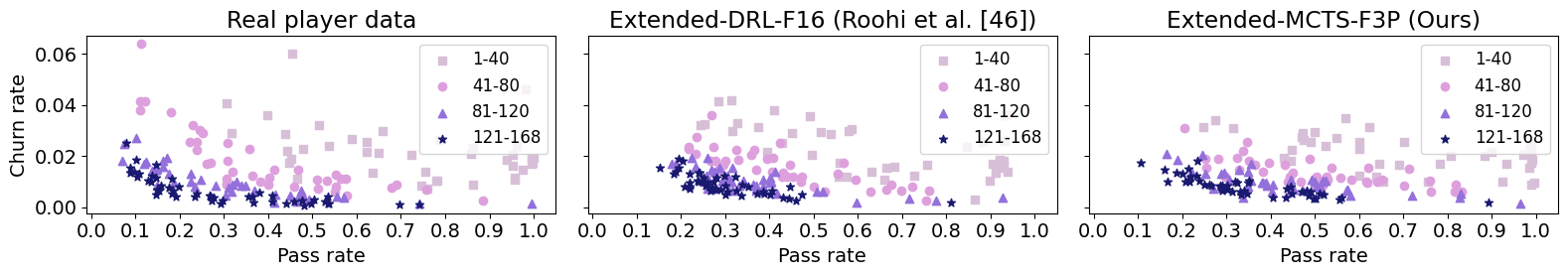}
	\caption{Human and predicted pass and churn rates over four groups of 168 consecutive game levels, comparing Roohi et al.'s \cite{roohi2020predicting} best original configuration (Extended-DRL-F16) approach to our best-performing extension (Extended-MCTS-F3P). The colors and shapes correspond to level numbers. Our predicted pass rates are slightly less truncated at the lower end.}	
	\label{fig:3_plots_ours}
\end{figure}

Table \ref{tab:mseTab} shows the mean squared errors (MSEs) of pass and churn rate predictions computed using the compared methods and 5-fold cross validation on the ground truth data from Roohi et al. \cite{roohi2020predicting}. Figure \ref{fig:passrate_mses_boxplots}  illustrates the same using box plots. The Baseline-DRL-F16 and Extended-DRL-F16 combinations are the original approaches analyzed by Roohi et al. For each fold, the extended predictor's errors were averaged over five predictions with different random seeds to mitigate the random variation caused by the predictor's stochastic optimization of the simulated player population parameters. We find that our Extended-MCTS-F3P configuration produces the overall most accurate predictions, and the proposed F3P features consistently outperform the other features in pass rate predictions.

\begin{table}[t!]
\centering
\caption{Means and standard deviations of pass and churn rate predictions mean squared errors, computed from all 12 compared configurations using 5-fold cross validation. The best results are highlighted using boldface. The Baseline-DRL-16 and Extended-DRL-16 configurations correspond to the original approaches developed and compared in Roohi et al. \cite{roohi2020predicting}.}~\label{tab:mseTab}
\begin{tabular}{|c|c|c|c|c|}
\hline
Predictor type & Agent&Features& Pass rate MSE & Churn rate MSE \\ \hline
Baseline&DRL&F16 & \makecell{$\mu=0.01890$, $\sigma=0.00514$}            & \makecell{$\mu=0.00014$, $\sigma=0.00003$} \\ \cline{3-5}
&&F3 & \makecell{$\mu=0.02041$, $\sigma=0.00453$}            & \makecell{$\mu=0.00012$, $\sigma=0.00003$} \\ \cline{3-5}
&&F3P & \makecell{$\mu=0.01742$, $\sigma=0.00368$}            & \makecell{$\mu=0.00012$, $\sigma=0.00003$} \\ \cline{2-5}
&MCTS&F16 & \makecell{$\mu=0.02040$, $\sigma=0.00370$}            & \makecell{$\mu=0.00013$, $\sigma=0.00003$} \\ \cline{3-5}
&&F3 & \makecell{$\mu=0.01919$, $\sigma=0.00305$}            & \makecell{$\mu=0.00012$, $\sigma=0.00003$} \\ \cline{3-5}
&&F3P & \makecell{$\mu=0.01452$, $\sigma=0.00207$}            & \makecell{$\mu=0.00012$, $\sigma=0.00003$} \\ \hline
Extended&DRL&F16 & \makecell{$\mu=0.01953$, $\sigma=0.00482$}            & \makecell{$\mu=0.00008$, $\sigma=0.00002$} \\ \cline{3-5}
&&F3 & \makecell{$\mu=0.01984$, $\sigma=0.00548$}            & \makecell{$\mu=0.00008$, $\sigma=0.00002$} \\ \cline{3-5}
&&F3P & \makecell{$\mu=0.01755$, $\sigma=0.00464$}        & \makecell{$\mu=0.00007$, $\sigma=0.00003$} \\ \cline{2-5}
&MCTS&F16 & \makecell{$\mu=0.02048$, $\sigma=0.00405$}            & \makecell{$\mu=0.00008$, $\sigma=0.00002$} \\ \cline{3-5}
&&F3 & \makecell{$\mu=0.01925$, $\sigma=0.00369$}            & \makecell{$\mu=0.00008$, $\sigma=0.00002$} \\ \cline{3-5}
&&F3P & \makecell{$\mu=\textbf{0.01419}$, $\sigma=\textbf{0.00216}$}        & \makecell{$\mu=\textbf{0.00007}$, $\sigma=\textbf{0.00002}$} \\ \hline
\end{tabular}
\vspace{-0.35cm}
\end{table}

Figure \ref{fig:3_plots_ours} shows scatter plots of churn and pass rates in all the 168 levels in our dataset. It affords a visual comparison between ground truth human data, the Extended-DRL-F16 configuration that corresponds to the best results in Roohi at al. \cite{roohi2020predicting}, and our best-performing Extended-MCTS-F3P configuration. Both simulated distributions appear roughly similar, but our Extended-MCTS-F3P yields slightly more realistic distribution of the later levels (121-168), with the predicted pass rates not as heavily truncated at the lower end as in Extended-DRL-F16.

\section{Discussion}
We have proposed and evaluated two extensions to the original player modeling approach of Roohi et al. \cite{roohi2020predicting}, in which difficulty and player engagement are operationalized as pass and churn rate, and predicted via features calculated from AI gameplay. We firstly used a game-playing agent based on DRL-enhanced MCTS instead of DRL only. We secondly reduced the set of prediction features based on correlation analysis with human ground truth data. We moreover calculated individual features based on different subsets of AI agent runs. 

Overall, \textit{both our proposed enhancements consistently improve prediction accuracy} over the original approach. The F3P features outperform the F16 and F3 ones irrespective of the agent or predictor, and the MCTS-F3P combinations outperform the corresponding DRL-F3P ones. The improvements are clearest in pass rate prediction: Our Extended-MCTS-F3P configuration improves the best of Roohi et al.'s \cite{roohi2020predicting} approaches by approximately one standard deviation, thus overcoming a crucial shortcoming of their work. 
We remark that the baseline churn rate predictions only slightly benefit from our enhancements, and the extended churn rate predictions are all roughly similar, no matter the choice of AI agent or features. We also replicate the results of Roohi et al. \cite{roohi2020predicting} in that the extended predictors with a population simulation outperform the baseline linear regression models.

A clear takeaway of this work is that computing feature averages from only a subset of the most successful game runs can improve AI-based player models. In other words, whenever there is a high variability in AI performance that does not match human players' ground truth, \textit{an AI agent's best-case behavior can be a stronger predictor of human play} than the agent's average behavior. As detailed below, these extensions must be tested in other types of games to evaluate their generalizability.

\section{Limitations and Future Work}
The amount of MCTS runs and the reward discount factor $\gamma$ have a considerable impact on the computing cost of our model. To deliver the best possible predictions at a low cost, it would be valuable to investigate in future work how changes to these parameters affect the accuracy of pass and churn rate predictions. 

While the evaluation of our approach on a commercially successful game demonstrates it applicability beyond mere research prototypes, it did not afford a comparison with existing AI playtesting approaches. We consider such a comparison of the prediction accuracy and computational costs on the same game an important next milestone. 

Our focus on a single game and data from only 168 levels moreover limits our findings with respect to the generalizability of our approach. For future work, it will hence be necessary to evaluate our models on larger datasets and several games from different genres.

Accurate player experience and behavior models can be used to assist game designers, or facilitate automated parameter tuning during design and runtime. For the future, we plan to test our method in the development process of a commercial game. More specifically, we aim to employ our algorithm to detect new game levels with inappropriate difficulty, and to investigate how changes to those levels would reduce player churn rate before presenting the levels to human playtesters or the end user.

\section{Conclusion}\label{sec:conclusion}
Our work advances the state-of-the-art in automated playtesting for player modeling, i.e., the prediction of human player behavior and experience using AI game-playing agents. Specifically, we have proposed and evaluated two enhancements to the recent approach by Roohi et al. \cite{roohi2020predicting} for predicting human player pass and churn rates as operationalizations of difficulty and engagement on 168 levels of the highly successful, free-to-play match-3 mobile game Angry Birds Dream Blast \citep{DB}. The required ground truth data can be obtained without refraining to obtrusive methods like questionnaires. 

By comparing against the original DRL-based approach of Roohi et al., we demonstrate that combining MCTS and DRL can improve the prediction accuracy, while requiring much less computation in solving hard levels than MCTS alone. Although combinations of MCTS and DRL have been utilized in many state-of-the-art game-playing systems, our work is the first to measure its benefits in predicting ground-truth human player data. Additionally, we propose an enhanced feature selection strategy that consistently outperforms Roohi et al.'s original features.

Moreover, in evaluating this feature selection strategy, we replicate Kristensen et al.'s \cite{kristensen2020estimating} findings that an  AI agent's best-case performance can be a stronger predictor of human player data than the agent's average performance. This shows in our extended predictor with stochastic population simulation, but has also been verified by computing the correlations between the human ground truth data and AI gameplay variables. This finding is valuable, as it should be directly applicable in other AI playtesting systems by running an agent multiple times and computing predictor features from a subset of best runs. We hope that our findings and improvements inspire future research, and foster  the application of automated player modelling approaches in games industry.

\section{Acknowledgments}\label{sec:acknowledge}
We thank the anonymous reviewers for their thorough and constructive comments. CG is funded by the Academy of Finland Flagship programme ``Finnish Center for Artificial Intelligence'' (FCAI). AR, HH and JT are employed by Rovio Entertainment. 

\bibliographystyle{ACM-Reference-Format}
\bibliography{references}

\received{February 2021}
\received[revised]{September 2021}
\received[accepted]{July 2021} 

\end{document}